\documentclass[a4paper]{spie}  %>>> use this instead for A4 paper
\usepackage[]{graphicx,epsfig,url}

\title{Automated classification method of COVID-19 cases from chest CT volumes using 2D and 3D hybrid CNN for anisotropic volumes}

\author{Masahiro ODA\supit{a,b}, Tong ZHENG\supit{b}, Yuichiro HAYASHI\supit{b}, Yoshito OTAKE\supit{c,d}, \\ Masahiro HASHIMOTO\supit{e}, Toshiaki AKASHI\supit{f}, Shigeki AOKI\supit{f}, and Kensaku MORI\supit{b,a,d}
\skiplinehalf
\supit{a}Information Strategy Office, Information and Communications, Nagoya University, \\
Furo-cho, Chikusa-ku, Nagoya, Aichi, 464-8601, Japan, \\
\supit{b}Graduate School of Informatics, Nagoya University, Nagoya, Japan, \\%Furo-cho, Chikusa-ku, Nagoya, Aichi, 464-8601, Japan; \\
\supit{c}Graduate School of Science and Technology, Nara Institute of Science and Technology, Nara, Japan, \\
\supit{d}Research Center for Medical Bigdata, National Institute of Informatics, Tokyo, Japan, \\
\supit{e}Department of Radiology, Keio University School of Medicine, Tokyo, Japan, \\
\supit{f}Department of Radiology, Juntendo University, Tokyo, Japan
}

\authorinfo{Further author information: 
(Send correspondence to M. Oda)\\
M. Oda: E-mail: moda@mori.m.is.nagoya-u.ac.jp, Telephone: +81 (0)52 789 5688\\
K. Mori: E-mail: kensaku@is.nagoya-u.ac.jp, Telephone: +81 (0)52 789 5689
}
\begin{document} 
  \maketitle 

%%%%%%%%%%%%%%%%%%%%%%%%%%%%%%%%%%%%%%%%%%%%%%%%%%%%%%%%%%%%% 
\begin{abstract}
This paper proposes an automated classification method of chest CT volumes based on likelihood of COVID-19 cases.
Novel coronavirus disease 2019 (COVID-19) spreads over the world, causing a large number of infected patients and deaths.
Sudden increase in the number of COVID-19 patients causes a manpower shortage in medical institutions.
Computer-aided diagnosis (CAD) system provides quick and quantitative diagnosis results.
CAD system for COVID-19 enables efficient diagnosis workflow and contributes to reduce such manpower shortage.
This paper proposes an automated classification method of chest CT volumes for COVID-19 diagnosis assistance.
We propose a COVID-19 classification convolutional neural network (CNN) that has a 2D/3D hybrid feature extraction flows.
The 2D/3D hybrid feature extraction flows are designed to effectively extract image features from anisotropic volumes such as chest CT volumes for diagnosis.
The flows extract image features on three mutually perpendicular planes in CT volumes and then combine the features to perform classification.
Classification accuracy of the proposed method was evaluated using a dataset that contains 1288 CT volumes.
An averaged classification accuracy was 83.3\%.
The accuracy was higher than that of a classification CNN which does not have 2D and 3D hybrid feature extraction flows.
\end{abstract}

%>>>> Include a list of keywords after the abstract 


%%%%%%%%%%%%%%%%%%%%%%%%%%%%%%%%%%%%%%%%%%%%%%%%%%%%%%%%%%%%%
\section{Introduction}

Novel coronavirus disease 2019 (COVID-19) was recognized in December 2019.
It spreads over the world causing the large number of infected patients.
The total numbers of cases and deaths related to COVID-19 are more than 424 million and 5 million in the world by February 21, 2022 \cite{worldmeters}.
Providing appropriate treatments to a patient and prevention of infection based on diagnosis result of the patient are important.
However, because of the rapid increase of COVID-19 patients, medical institutions are suffering from a manpower shortage.
Development of a computer aided diagnosis (CAD) system for COVID-19 is pressing demanded to reduce load on medical staffs.
Reverse transcriptase polymerase chain reaction testing (RT-PCR) is used to diagnose COVID-19.
However, the sensitivity of RT-PCR is not high, ranging from 42\% to 71\% \cite{Simpson20}.
In contrast, the sensitivity of chest CT image-based COVID-19 diagnosis is reported as 97\% \cite{Ai20}.
Chest CT is effective for diagnosis of viral pneumonia including COVID-19.
CT image-base CAD systems are important in COVID-19 diagnosis.
To develop a CAD system for COVID-19, automated classification method of CT image/volume is necessary.

Many automated classification method of COVID-19 patients using chest CT images/volumes have been proposed.
Deep learning-based methods are commonly used to classify CT images.
Classification methods using 2D CNN models \cite{Anwar20,Islam20,James20,Tabarisaadi20}, 3D CNN models \cite{Xue21}, weakly supervised method \cite{Hu20}, uncertainty of CNN model prediction \cite{Shamsi21}, and federated learning \cite{Dou21}.
In such methods, CNN classifiers are used as fundamental components.
Improving accuracy of COVID-19 classification CNN helps achieving better performance for such methods.

We propose an automated classification method of chest CT volumes based on likelihood of COVID-19 cases.
We perform classification using a COVID-19 classification convolutional neural network (CNN) that has a 2D/3D hybrid feature extraction flows.
Chest CT volumes of various slice thickness (1.0 to 5.0 mm) are used in diagnosis.
The variation of slice thickness makes classification difficult using 3D CNNs that have convolution kernels of isotropic sizes such as $3 \times 3 \times 3$ or $5 \times 5 \times 5$.
To tackle this problem, we propose the 2D/3D hybrid feature extraction flows that is effective for classifying anisotropic volumes such as chest CT volumes.
The 2D/3D hybrid feature extraction flows extract image features on multiple 2D planes including the axial, coronal, and sagittal planes in CT volumes separately.
Then, extracted 2D features are combined to perform 3D image classification.
To perform 3D image classification in the CNN, dense pooling connections and dilated convolutions are introduced to the CNN to utilize both local and global spatial image features.

%%%%%%%%%%%%%%%%%%%%%%%%%%%%%%%%%%%%%%%%%%%%%%%%%%%%%%%%%%%%%
\section{Method} \label{sec:method}

\subsection{Overview}

\begin{figure}[tb]
\begin{center}
\includegraphics[width=0.95\textwidth]{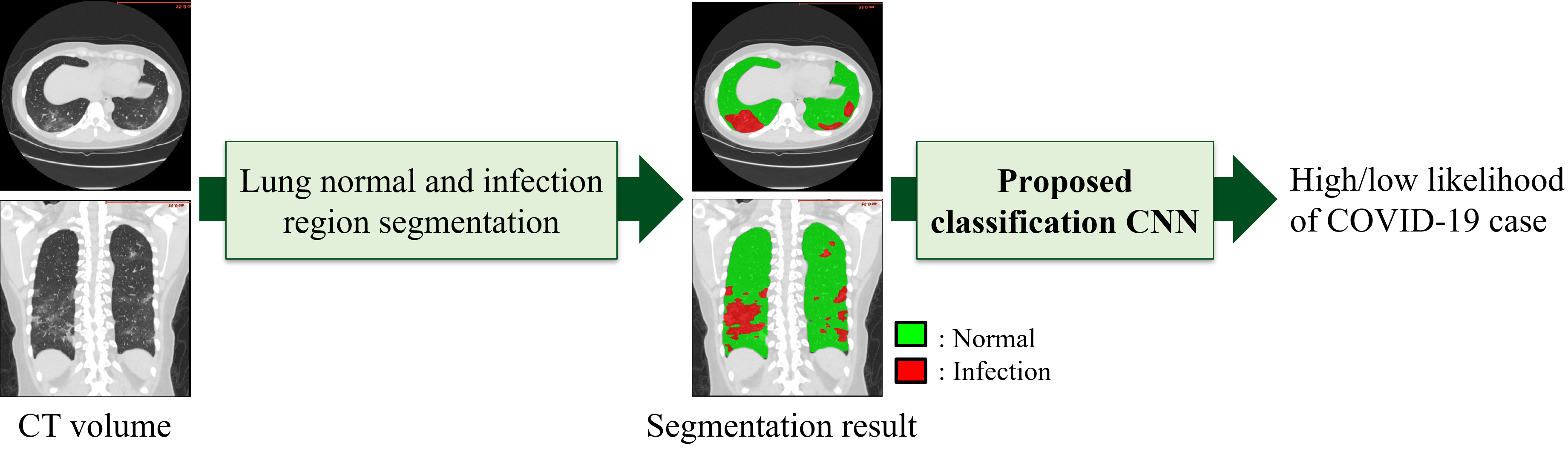}
\caption{Processing flow of automated COVID-19 classification from chest CT volume. Lung regions including normal and infection regions are automatically segmented by previously proposed method \cite{Oda21}. We classify volume into high or low likelihoods of COVID-19 cases automatically using proposed COVID-19 classification CNN that has a 2D/3D hybrid feature extraction flows.}
\label{fig:processflow}
\end{center}
\end{figure}

The proposed method classifies a chest CT volume into two classes that correspond to high or low likelihoods of COVID-19 cases.
The likelihood is defined based on CT image findings confirmed by radiologists.
Processing flow of the proposed method is summarized in Fig. \ref{fig:processflow}.

As a pre-processing of the classification, we generate a lung volume using an automated lung region segmentation method from a CT volume.
We apply the COVID-19 classification CNN to the lung volume to perform classification.
Two class classification result is obtained from the CNN.

\subsection{Lung volume generation}

We automatically segment lung regions from a chest CT volume using a segmentation method \cite{Oda21}.
The segmentation method outputs normal and infection regions in the lung.
The infection regions may contain ground glass opacities and consolidations.
The two regions are combined to make lung regions.
We make a sub-volume by clipping a region where is covered by the lung regions from the chest CT volume.
The sub-volume is scaled to $144 \times 144 \times 144$ voxels.
The scaled sub-volume is called as a lung volume.

\subsection{Lung volume classification by COVID-19 classification CNN with 2D/3D hybrid feature extraction flows}

\begin{figure}[tb]
\begin{center}
\includegraphics[width=0.95\textwidth]{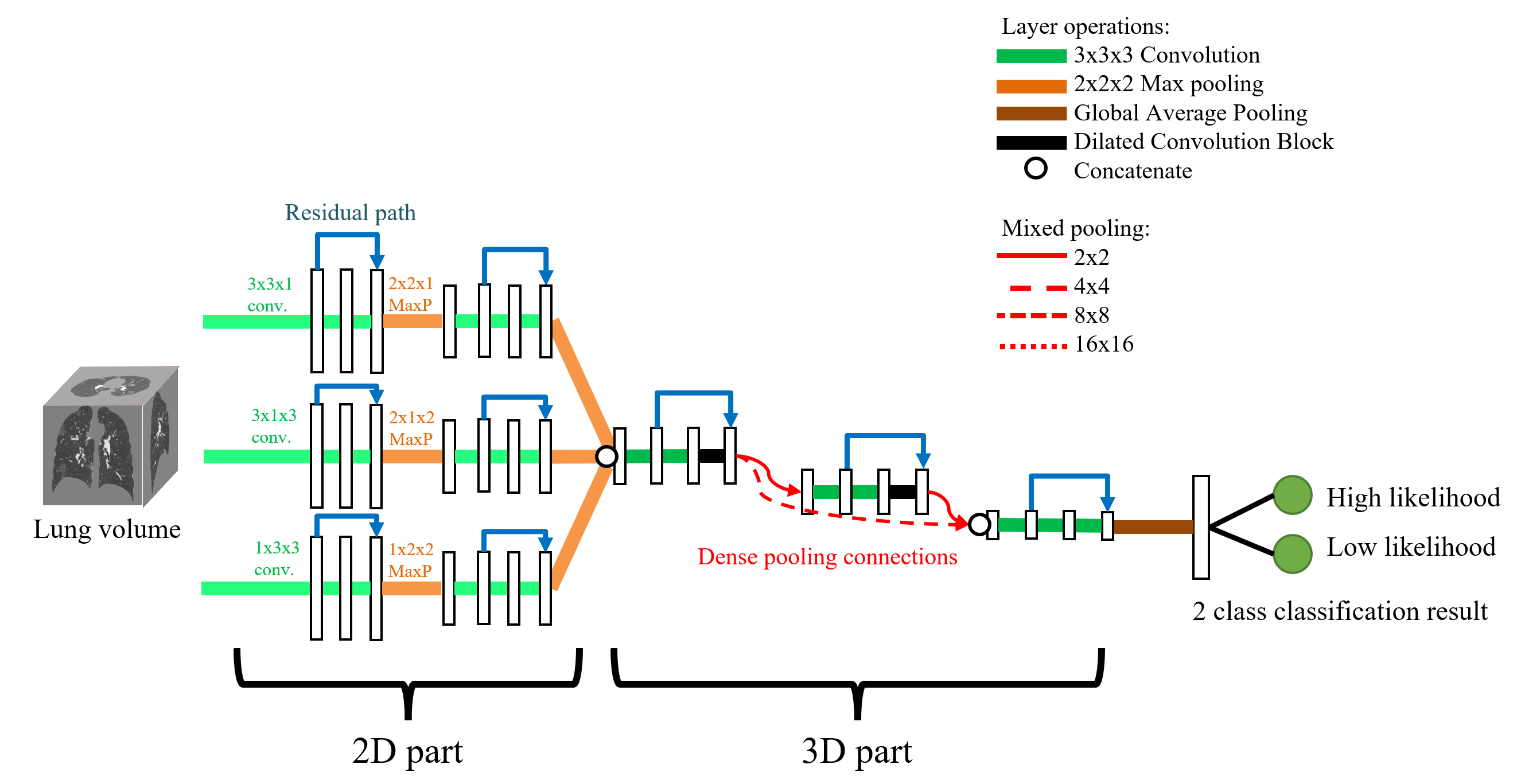}
\caption{Structure of proposed COVID-19 classification CNN with 2D/3D hybrid feature extraction flows. Input of CNN is lung volume. Lung volume is processed by three parallel process flows in 2D part. Three flows in 2D part perform 2D image feature extraction on axial, coronal, and sagittal planes. Extracted features are combined and then processed in 3D part. In 3D part, dilated convolutions and dense pooling connections are used to effectively extract spatial features.}
\label{fig:cnnstructure_propose}
\end{center}
\end{figure}

We classify the lung volume into two classes that correspond to high and low likelihoods of COVID-19 cases using the COVID-19 classification CNN.
The structure of the CNN is shown in Fig. \ref{fig:cnnstructure_propose}.
The CNN is trained using a training dataset and then classification is performed using a testing dataset.

The CNN has 2D/3D hybrid feature extraction flows to extract image features effectively from anisotropic volumes.
The first half of the CNN is a 2D part.
Most of chest CT volumes have anisotropic resolutions.
Convolution kernels of isotropic sizes are not suitable for image feature extraction from anisotropic volumes.
We use convolution kernels of sizes $3 \times 3 \times 1$, $3 \times 1 \times 3$, and $1 \times 3 \times 3$ for image feature extractions on the axial, coronal, and sagittal planes, respectively, in the 2D part of the 2D/3D hybrid feature extraction flows.
In this 2D part, 2D image processing-based image feature extractions on the three planes are performed in parallel to the volume.
The three 2D-based features are then combined using a concatenate operation.

The combined features are processed by the last half of the CNN.
The last half of the CNN is a 3D part.
We apply 3D convolutions and poolings to the features.
To extract both local and global spatial image features, we use dilated convolutions and dense pooling connections in the 3D part.
Dilated convolution \cite{Yu16} was proposed to utilize sparsely-distributed features in convolution operations.
It performs a sparse convolution operation from feature maps.
Dense pooling connection \cite{playout18} was proposed to utilize spatial information of multiple scales in CNNs and FCNs.
Mixed poolings \cite{playout18} are used in the dense pooling connections instead of max poolings to reduce information loss by max pooling operations.
The mixed pooling is implemented as a combination of max and average poolings.

The last layer of the CNN outputs two class classification result.

%\begin{figure}[tb]
%\begin{center}
%%\begin{tabular}{c}
%\includegraphics[width=0.8\textwidth, clip, trim=0 140 0 0]{fig/network.eps}
%%\end{tabular}
%\end{center}
%\caption{FCN for lung infection and normal region segmentation. White boxes are feature maps or images. Numbers below boxes are numbers of kernel or color channel. Dense pooling connections are represented as red connections, which are implemented as combination of mixed poolings. Two dilated convolution blocks are inserted to encoding path. Structure of dilated convolution block is shown in box.}
%\label{fig:network}
%\end{figure}

%%%%%%%%%%%%%%%%%%%%%%%%%%%%%%%%%%%%%%%%%%%%%%%%%%%%%%%%%%%%%
\section{Experiments and Results} \label{sec:results}

We applied the proposed method to 1288 chest CT volumes including COVID-19 cases and non-COVID-19 cases.
The CT volumes were taken in multiple medical institutions in Japan.
%Specifications of the CT volumes are: image size was 512$\times$512 pixels, slice number was 56 to 722, pixel spacing was 0.63 to 0.78 mm, and slice thickness was 1.00 to 5.00 mm.
The ground truth class labels of CT volumes were given by radiologists.
We used 1030 cases for training and 258 cases for testing.
Separations of training and testing cases were randomly performed.
In the training of the CNN, the number of minibatch was 2 and the training epoch number was 50.
We applied a data augmentation process to the training CT volumes to improve generalization ability of the CNN.
The data augmentation process includes 3D shift, rotation, and elastic deformation with random parameters.

\begin{figure}[tb]
\begin{center}
\includegraphics[width=0.95\textwidth]{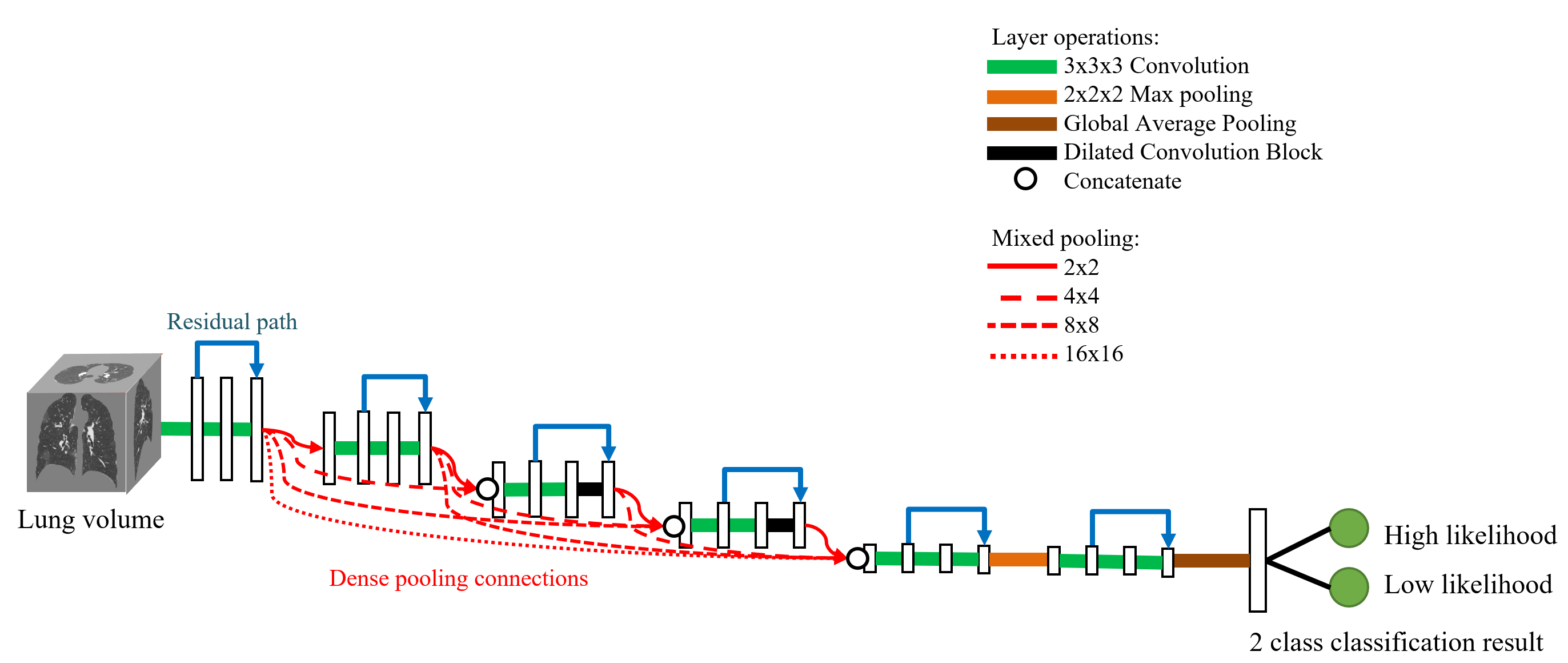}
\caption{Structure of classification CNN that does not have 2D part of 2D/3D hybrid feature extraction flows (CNN w/o 2D/3D hybrid).}
\label{fig:cnnstructure_3d}
\end{center}
\end{figure}

To evaluate effectiveness of the 2D/3D hybrid feature extraction flows, we made a classification CNN that does not have 2D part (CNN w/o 2D/3D hybrid).
The structure of the CNN w/o 2D/3D hybrid is shown in Fig. \ref{fig:cnnstructure_3d}.
The CNN w/o 2D/3D hybrid was made by replacing 2D part in the COVID-19 classification CNN with 3D convolutions and dense pooling connections.

\begin{figure}[tb]
\begin{center}
\includegraphics[width=0.8\textwidth]{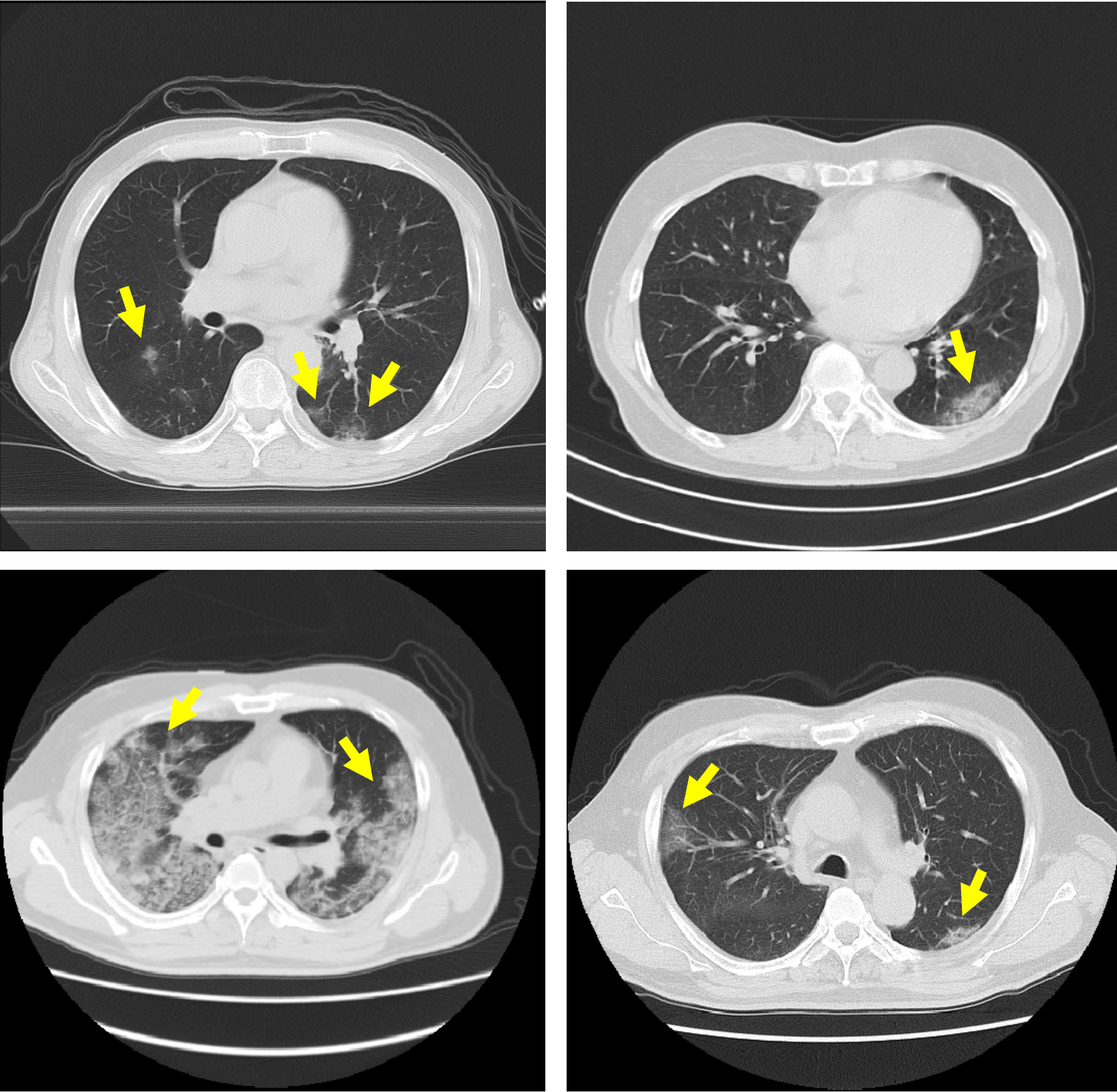}
\caption{Axial slices of CT volumes that were identified as high likelihood of COVID-19 cases by radiologists. These cases were correctly classified into high likelihood class by proposed CNN. Infection regions are indicated by arrows. Infection regions have large variations in their sizes and CT values. Proposed CNN correctly classified such cases considering characteristics of infection regions caused by COVID-19.}
\label{fig:result_high}
\end{center}
\end{figure}

\begin{figure}[tb]
\begin{center}
\includegraphics[width=0.8\textwidth]{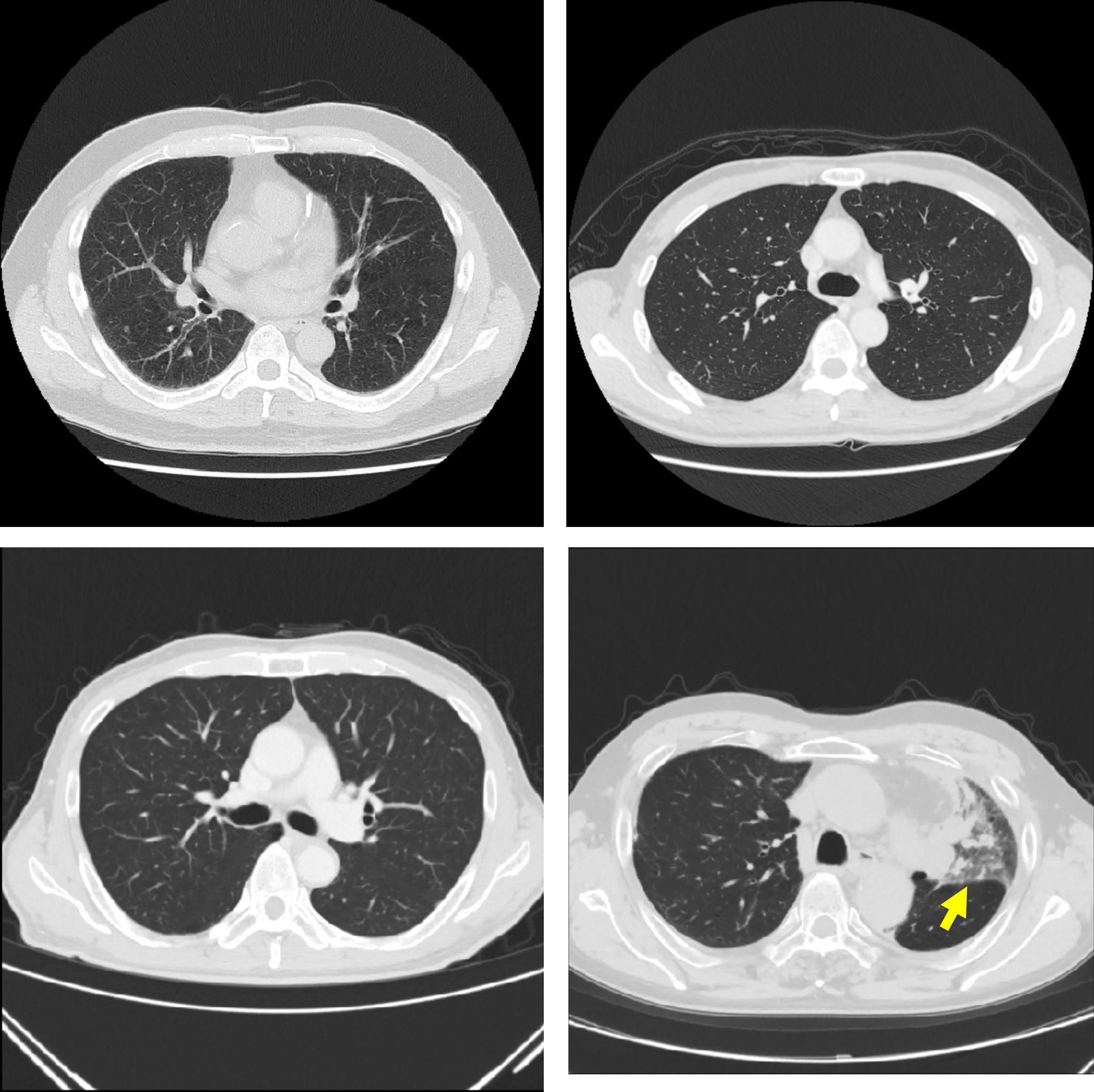}
\caption{Axial slices of CT volumes that were identified as low likelihood of COVID-19 cases by radiologists. These cases were correctly classified into the low likelihood class by the proposed CNN. Infection region is indicated by arrow. Some cases have infection regions that were caused by disorders other than COVID-19. Proposed CNN correctly classified such cases as low likelihood cases.}
\label{fig:result_low}
\end{center}
\end{figure}

We performed random training and testing data separation and CNN training four times.
Averaged classification accuracies of the proposed COVID-19 classification CNN and the CNN w/o 2D/3D hybrid were ${\bf 83.3\pm1.8\%}$ and 79.5$\pm$2.1\%.
The confusion matrix of classification result by the proposed COVID-19 classification CNN is shown in Table \ref{tab:confmatrix}.
Some cases that were correctly classified as high and low likelihood of COVID-19 cases were shown in Figs. \ref{fig:result_high} and \ref{fig:result_low}, respectively.

\begin{table}[tb]
\caption{Confusion matrix of classification result by proposed COVID-19 classification CNN when we obtained 84.1\% of classification accuracy. High and Low indicate high and low likelihoods of COVID-19 cases.}\label{tab:confmatrix}
\begin{center}
\begin{tabular}{|c|c|c|c|}
\hline
\multicolumn{2}{|c|}{} & \multicolumn{2}{|c|}{Estimation result}  \\ \cline{3-4}
\multicolumn{2}{|c|}{} & High & Low \\ \hline
Ground truth & High & 135 & 26 \\ \cline{2-4}
 & Low & 15 & 82 \\ \hline
\end{tabular}
\end{center}
\end{table}

The proposed COVID-19 classification CNN classified chest CT volumes with high accuracy.
The comparative experiment proofed the 2D/3D hybrid feature extraction flows contributes to improve classification accuracy.
The flows effectively extracts image features even from anisotropic volumes.
Use of the 2D image processing-based kernels reduces the number of trainable parameters in CNN.
We can obtain better training results even from limited number of training data.

%%%%%%%%%%%%%%%%%%%%%%%%%%%%%%%%%%%%%%%%%%%%%%%%%%%%%%%%%%%%%
\section{Discussion}

We propose a COVID-19 classification CNN that has novel 2D/3D hybrid feature extraction flows in this paper.
The 2D/3D hybrid feature extraction flows is quite effective in classification of anisotropic volume such as chest CT volumes for diagnosis.
By using the 2D/3D hybrid feature extraction flows, classification accuracies were improved from 79.5\% to 83.3\% on averages.

As shown in Fig. \ref{fig:result_high} that shows cases of high likelihood of COVID-19 cases, infection regions in the lung have large variations in their sizes and CT values.
Some cases have large infection regions and some cases have small infection regions.
Proposed CNN correctly classified such cases considering characteristics of infection regions caused by COVID-19.
Figure \ref{fig:result_low} shows cases of low likelihood of COVID-19 cases.
They were correctly classified into the low likelihood class by the proposed CNN.
Most of such cases have no obvious infection regions.
Some cases have infection regions that were caused by disorders other than COVID-19.
The proposed CNN correctly distinguished them based on image features.

The layers in the 2D part of the proposed CNN is implemented by limiting kernel sizes of 3D convolutions and 3D pooling layers.
Such simple implementation can be easily used with many existing deep learning frameworks.
Furthermore, the 2D/3D hybrid feature extraction flows can be used to effectively extract features from anisotropic volumes, not only from CT volumes but also other types of volumetric images.
The proposed method has a high potential to improve performances of volumetric image analysis.

%%%%%%%%%%%%%%%%%%%%%%%%%%%%%%%%%%%%%%%%%%%%%%%%%%%%%%%%%%%%%
\section{Conclusions}

We proposed a COVID-19 classification CNN for diagnosis assistance using chest CT volumes.
Because most of chest CT volumes have anisotropic resolutions, we introduced the 2D/3D hybrid feature extraction flows in the CNN.
In the flow, three parallel 2D feature extractions on the axial, coronal, and sagittal planes are performed.
The feature extraction results are then combined and processed by the 3D processing part in the CNN.
Dilated convolutions and dense pooling connections are introduced in the CNN to effectively extract spatial features.
In the evaluation using 1288 chest CT volumes, the proposed COVID-19 classification CNN achieved classification accuracy of 83.3$\pm$1.8\%.
It was higher than the CNN w/o 2D/3D hybrid feature extraction flows.
Future work includes introduction of pre-processing that contributes for classification accuracy improvement and development of a COVID-19 CAD system.

\acknowledgments % equivalent to \section*{ACKNOWLEDGMENTS}       
 
Parts of this research were supported by the AMED Grant Numbers 18lk1010028s0401,
JP19lk1010036, and JP20lk1010036, the NICT Grant Number 222A03, the JST CREST Grant Number JPMJCR20D5, the MEXT/JSPS KAKENHI Grant Numbers 26108006, 17H00867, and 17K20099, the JSPS Bilateral International Collaboration Grants. We used the Japan Medical Image Database (J-MID) created by the Japan Radiological Society with support by the AMED Grant Number JP20lk1010025.

%%%%%%%%%%%%%%%%%%%%%%%%%%%%%%%%%%%%%%%%%%%%%%%%%%%%%%%%%%%%%
%%%%% References %%%%%

\bibliography{22spie_paper_cite}   %>>>> bibliography data in report.bib
\bibliographystyle{spiebib}   %>>>> makes bibtex use spiebib.bst

\end{document}